\documentclass{article}
\usepackage{PRIMEarxiv}

\usepackage[utf8]{inputenc} 
\usepackage[T1]{fontenc}    
\usepackage{hyperref}       
\usepackage{url}            
\usepackage{booktabs}       
\usepackage{amsfonts}       
\usepackage{nicefrac}       
\usepackage{microtype}      
\usepackage{titlesec}
\usepackage{graphicx}

\usepackage{array}
\usepackage{float}
\usepackage{subfig}
\usepackage{multirow}
\usepackage{rotating}
\usepackage{makecell}

\usepackage{amsmath}
\usepackage{algorithmic}
\usepackage{textcomp}
\usepackage{stfloats}
\usepackage{verbatim}
\usepackage{cite}
\usepackage{pifont}
  
\title{Hierarchical Self-Supervised Representation Learning for Depression Detection from Speech
}

\author{
  Yuxin Li \\
  College of Computing and Data Science \\
  Nanyang Technological University \\
  Singapore\\
  \texttt{yuxin.li@ntu.edu.sg} \\
  \And
  Eng Siong Chng \\
  College of Computing and Data Science \\
  Nanyang Technological University \\
  Singapore\\
  \texttt{aseschng@ntu.edu.sg} \\
  \And
  Cuntai Guan \\
  College of Computing and Data Science \\
  Center of AI in Medicine \\
  Nanyang Technological University \\
  Singapore\\
  \texttt{ctguan@ntu.edu.sg} \\
}

\begin{document}
\maketitle

\begin{abstract}
Speech-based depression detection (SDD) has emerged as a non-invasive and scalable alternative to conventional clinical assessments. However, existing methods still struggle to capture robust depression-related speech characteristics, which are sparse and heterogeneous. Although pretrained self-supervised learning (SSL) models provide rich representations, most recent SDD studies extract features from a single layer of the pretrained SSL model for the downstream classifier. This practice overlooks the complementary roles of low-level acoustic features and high-level semantic information inherently encoded in different SSL model layers. To explicitly model interactions between acoustic and semantic representations within an utterance, we propose a hierarchical adaptive representation encoder with prior knowledge that disengages and re-aligns acoustic and semantic information through asymmetric cross-attention, enabling fine-grained acoustic patterns to be interpreted in semantic context. In addition, a Connectionist Temporal Classification (CTC) objective is applied as auxiliary supervision to handle the irregular temporal distribution of depressive characteristics without requiring frame-level annotations. Experiments on DAIC-WOZ and MODMA demonstrate that HAREN-CTC consistently outperforms existing methods under both performance upper-bound evaluation and generalization evaluation settings, achieving Macro F1 scores of 0.81 and 0.82 respectively in upper-bound evaluation, and maintaining superior performance with statistically significant improvements in precision and AUC under rigorous cross-validation. These findings suggest that modeling hierarchical acoustic–semantic interactions better reflects how depressive characteristics manifest in natural speech, enabling scalable and objective depression assessment.
\end{abstract}


\section{Introduction}
Depression is a prevalent and debilitating mental health disorder, affecting over 280 million individuals worldwide \cite{who_depression}. It is often underdiagnosed and undertreated due to the reliance on subjective self-reports and the limited availability of trained mental health professionals \cite{shi2021research}. These challenges have motivated the development of automated, objective, and scalable approaches for depression assessment \cite{kwon2021depression}.

Speech-based depression detection (SDD) has emerged as a promising alternative by leveraging vocal biomarkers that naturally reflect an individual’s emotional and cognitive state. Prior studies have reported that depression is associated with characteristic speech patterns, such as reduced pitch variability, slower articulation rate, longer pauses, and increased negative semantic content \cite{cummins2015review, low2010detection}. Despite these findings, depression-related speech characteristics are often subtle, heterogeneous, and temporally sparse, making them difficult to capture effective depressive indicators.

Recent advances in self-supervised learning (SSL) have substantially improved speech representation learning by enabling models to learn hierarchical and contextualized representations from large-scale unlabeled audio data. SSL models such as Wav2Vec~2.0 \cite{baevski2020wav2vec}, HuBERT \cite{hsu2021hubert}, WavLM \cite{chen2022wavlm}, and Whisper \cite{radford2023robust} encode diverse information across layers, ranging from low-level acoustic patterns to high-level semantic and speaker-related attributes \cite{chen2022wavlm}. However, in the context of SDD, most existing methods rely on representations extracted from a single SSL layer, or selected heuristically or through ablation studies \cite{Toto2021, lin2022deep, wu2023self, zhang2024improving, chen2022speechformer, wang2024speechformer}. This design treats SSL layers as interchangeable features and overlooks the complementary and hierarchical nature of acoustic and semantic information encoded across different layers.

Importantly, many depression-related characteristics do not reside exclusively in either low-level acoustics or high-level semantics, but instead emerge from their interaction \cite{williamson2016detecting, wu2024depression}. Fine-grained acoustic patterns, such as prosodic variability or speech fluency, often need to be interpreted within semantic and contextual information to become informative for depression detection \cite{yang2012detecting}. This observation suggests that simple feature selection or indiscriminate fusion of SSL layers is insufficient \cite{yue2024hierarchical}, and that a principled hierarchical modeling approach is required to explicitly capture cross-level interactions.

Motivated by this gap, we propose \textit{HAREN}-CTC, a hierarchical adaptive representation encoder for depression-related features built upon self-supervised speech models, with an explicit inductive bias for aligning acoustic and semantic information. Rather than introducing another generic feature fusion mechanism, \textit{HAREN}-CTC systematically structures SSL representations into shallow acoustic-focused and deep semantic-rich subspaces, and models their interactions asymmetrically. This design enables fine-grained acoustic cues to be interpreted under a semantic context, which is critical for capturing subtle and heterogeneous depressive speech patterns.

At the core of the proposed framework is the Hierarchical Adaptive Representation Encoder (\textit{HAREN}), which consists of two tightly coupled components. First, the Hierarchical Adaptive Soft Grouping module disentangles multi-layer SSL representations into acoustic and semantic subspaces. Second, Semantic-conditioned Cross-Attention Fusion employs a directed cross-attention mechanism that treats deep semantic representations as queries and shallow acoustic representations as keys and values, allowing the model to identify depression-relevant acoustic patterns conditioned on semantic context rather than through symmetric or indiscriminate fusion.

For the optimization objective, we further extend the utterance-level depression detection objective by incorporating an auxiliary Connectionist Temporal Classification (CTC) loss \cite{graves2012connectionist, wang2024speechformer} to address the sparsity and irregular temporal distribution of depression-related characteristics. This CTC objective serves as temporal structure regularization, providing weak alignment signals without requiring frame-level annotations and encouraging the model to capture latent temporal structure in depressive speech.

We evaluate the proposed framework on two benchmark datasets, DAIC-WOZ and MODMA, under both upper-bound and generalization evaluation settings.

The remainder of this paper is organized as follows. Section II reviews related work on speech-based depression detection, ranging from traditional approaches to deep learning-based methods. Section III describes the methodology including the structure of the proposed \textit{HAREN}-CTC framework and its training objectives. Section IV outlines the experimental setup, including datasets, preprocessing, implementation details, baseline downstream depression detection architectures, and two evaluation protocols. Section V presents the experimental results and discusses the key findings. Finally, Section VI concludes the paper.

\section{RELATED WORK}

\subsection{Traditional Depression Detection from Speech}
Early research on depression detection from speech primarily relied on handcrafted acoustic features combined with traditional machine learning models. Commonly used features include Mel-Frequency Cepstral Coefficients (MFCCs), zero-crossing rates and spectral entropy, which capture fundamental properties of speech signals. These features were then fed into classifiers such as Support Vector Machines (SVMs), Random Forests, and Logistic Regression models. \cite{rejaibi2022, prabhu2022harnessing,guo2022automatic,kwon2021depression,zhang2024multimodal,morales2018linguistically}. While these methods provided an initial foundation for automated SDD, they suffered from limited generalization capabilities. The reliance on predefined feature sets made them highly sensitive to speaker variations, linguistic differences, and environmental noise \cite{tasnim2022cost}. 

\subsection{Deep Learning Methods for Depression Detection}
With the rise of deep learning, speech-based depression detection moved beyond handcrafted features toward end-to-end learning. Early deep models like Convolutional Neural Networks (CNNs) and Recurrent Neural Networks (RNNs) effectively learned hierarchical speech features from raw audio/spectrograms, establishing foundations for modern SDD systems \cite{chlasta2019automated, saidi2020hybrid, vazquez2020automatic, muzammel2020audvowelconsnet, guo2022topic, ye2021multi, lam2019context, marriwala2023hybrid, lin2020towards, dubagunta2019learning}. To better model long-range dependencies and temporal variation, CNNs were often combined with Long Short-Term Memory (LSTM) networks or replaced with Transformer-based architectures, which provided stronger temporal modeling capacity \cite{zhao2020hierarchical, yin2023depression}.

Hybrid models such as DepAudioNet \cite{ma2016depaudionet} integrated CNNs and LSTMs to jointly learn spatial and temporal features, enabling more comprehensive detection of depression-related cues. These models laid the groundwork for many SDD systems and remain relevant today, often serving as downstream classifiers for more advanced feature encodings.

Despite their effectiveness, such models often struggle with generalization due to limited labeled datasets. Furthermore, they are highly sensitive to speaker variability, noise, and recording conditions. This motivated a shift toward richer, transferable representations such as those learned through self-supervised learning.

\subsection{Self-Supervised Representations and Their Integration}

Self-supervised learning (SSL) has significantly advanced speech representation learning by enabling models to learn from large-scale unlabeled audio. Models such as Wav2Vec 2.0 \cite{baevski2020wav2vec}, HuBERT \cite{hsu2021hubert}, WavLM \cite{chen2022wavlm}, and Whisper \cite{radford2023robust} produce multi-layer contextualized representations that encode both low-level acoustic details and high-level semantic content. These representations are highly transferable and have been widely adopted as input features for downstream depression detection tasks, typically by selecting features from a single SSL layer \cite{Zhang2021Depa, Toto2021, lin2022deep, chen2022speechformer, ravi2022step, wu2023self, zhang2024improving, SFTN2024, wang2024speechformer}.

Several studies have demonstrated the potential of SSL in SDD. For instance,  Wu et al. \cite{wu2023self} conducted a comparative evaluation of Wav2Vec 2.0, HuBERT, and WavLM across different layers and found significant variation in performance depending on which layer was used, highlighting the importance of layer selection but without explicitly modeling interactions between representation levels. SpeechFormer \cite{chen2022speechformer} proposed a novel transformer-based architecture tailored for depression detection from speech. The model incorporates learnable temporal downsampling modules that dynamically adjust the resolution of acoustic features across different transformer layers. This allows the network to capture fine-grained local patterns and long-range temporal dependencies in speech signals, primarily through temporal modeling within a single representation stream. Specifically, the model takes Wav2Vec 2.0 features as input and processes them through a stack of transformer encoder layers, each preceded by a learnable downsampling block that reduces the sequence length while preserving information. Additionally, a temporal attention module is applied at the final layer to focus on segments that are more emotionally salient or depression-relevant. The final segment-level predictions are aggregated to obtain utterance-level or session-level classification results. This method achieves an average F1 of 0.694 on DAIC-WOZ. More recently, SpeechFormer-CTC \cite{wang2024speechformer} extended this approach by integrating WavLM representations with a CTC-based loss to align segment-level acoustic inputs with coarse depression labels. By incorporating temporal alignment constraints and multi-task learning with auxiliary depression classification, the model achieved a state-of-the-art F1 score on DAIC-WOZ. However, its design focuses on temporal alignment within a single SSL feature stream rather than explicit hierarchical representation interaction.

However, most existing models using SSL representations adopt a single-layer feature, either selected heuristically or via ablation. This design overlooks the hierarchical structure of SSL models, where different layers specialize in distinct types of information: Shallow layers primarily capture detailed prosodic and phonetic information, whereas deep layers encode more abstract semantic and speaker-level characteristics \cite{chen2022wavlm, de2024layer, ashihara2024self}. The potential of explicitly structuring and aligning multiple SSL representation layers to model their complementary interactions remains underexplored in SDD.

\begin{figure*}[thp]
    \centering
    \includegraphics[width=1\linewidth]{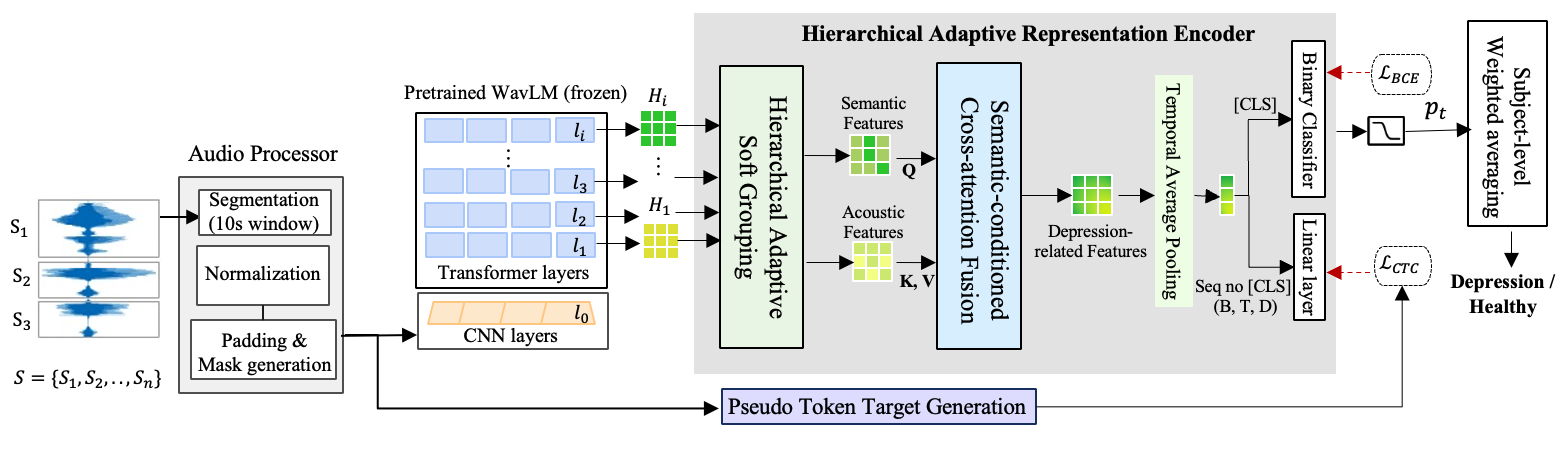}
    \caption{Overview of the \textit{HAREN}-CTC architecture. The model hierarchically structures acoustic and semantic SSL representations, aligns them through semantic-conditioned cross-attention, and incorporates CTC-based auxiliary supervision with HuBERT pseudo-labels to handle temporally sparse depression-related cues.}
    \label{fig:HAREN}
\end{figure*}

\section{METHODOLOGY}

\subsection{Model Overall}

We propose \textit{HAREN}-CTC, a hierarchical representation learning framework for detecting depression in speech. \textit{HAREN}-CTC centers on a Hierarchical Adaptive Representation Encoder which structures and fuses acoustic and semantic information across multiple abstraction levels. During training, an auxiliary CTC-based temporal regulation branch is introduced to encourage event-level temporal consistency in depression-related representations. An overview of the architecture and training objectives is shown in Fig.~\ref{fig:HAREN}.

\subsection{Problem Definition}
We formulate depression detection as a hierarchical binary classification task. Given a subject's session $S = \{S_1, S_2, \ldots, S_n\}$ containing $n$ utterances, we first perform utterance-level classification. For each utterance $S_i$, the model predicts $p_i \in [0,1]$ indicating the probability of depression state for that utterance. At the subject level, we aggregate predictions across all utterances belonging to a subject through weighted averaging. The final classification is determined by thresholding, where a subject is classified as depressed if $p_{subject} > 0.5$, and non-depressed otherwise.

\subsection{Hierarchical Adaptive Representation Encoder}
\subsubsection{Hierarchical Adaptive Soft Grouping}
\begin{figure}
    \centering
    \includegraphics[width=0.5\linewidth]{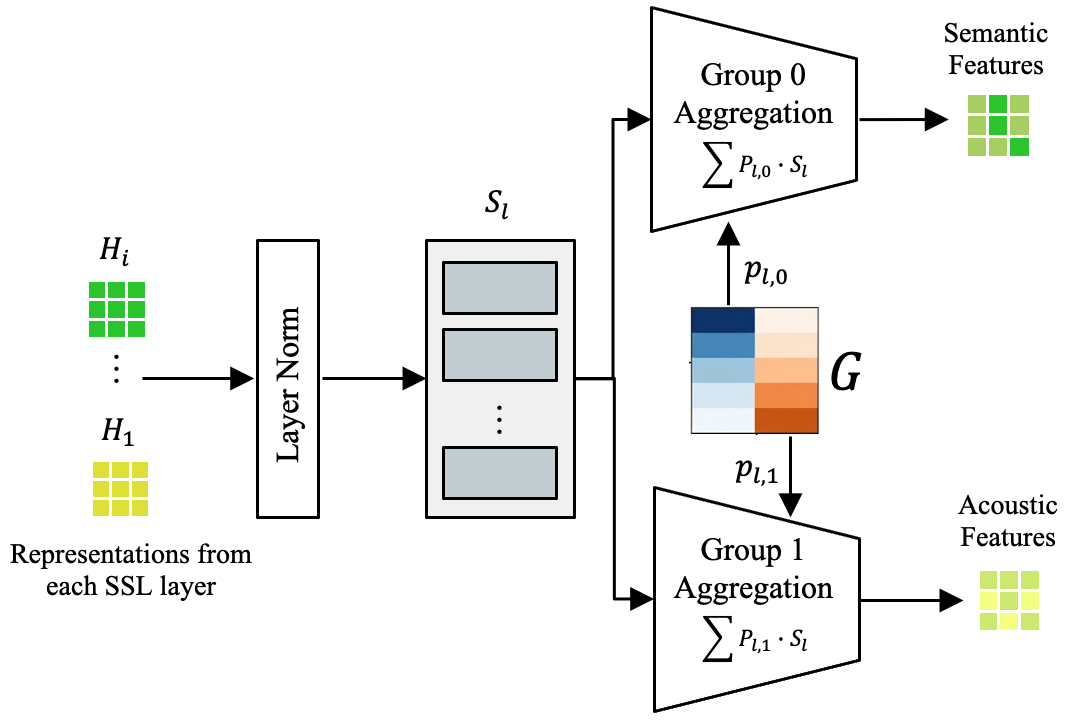}
    \caption{The Mechanism of Hierarchical Adaptive Soft Grouping}
    \label{fig:grouping}
\end{figure}

We first perform Hierarchical Adaptive Soft Grouping to explicitly disentangle multi-layer SSL representations into acoustic and semantic subspaces before modeling their interactions. As shown in Fig \ref{fig:grouping}, let $\{H_{i_1}, \ldots, H_{i_m}\}$ denote representations from each SSL layer. We introduce a learnable assignment matrix $G \in \mathbb{R}^{m \times 2}$ to partition these representations into shallow and deep subspaces. Each row of $G$ corresponds to one SSL layer, while the two columns represent the acoustic and semantic subspaces, respectively.

Given the prior knowledge that shallow layers of SSL models encode more acoustic and phonemic information while deeper layers capture more semantic information, we initialize $G$ using an exponential decay strategy that biases earlier layers toward the acoustic subspace and deeper layers toward the semantic subspace. This initialization serves only as a soft prior; the layer-to-subspace assignments remain fully learnable during training. Each row of $G$ is normalized using a softmax operation:
\begin{equation}
P_{l,k} = \frac{\exp(G_{l,k})}{\sum_{k'=1}^{2} \exp(G_{l,k'})},
\end{equation}

which yields adaptive probabilities for assigning layer $l$ to subspace $k$. The resulting structured representations are obtained via weighted aggregation:
\begin{equation}
U^{(k)} = \sum_{l=1}^{m} P_{l,k} H_l, \quad k \in \{\text{shallow}, \text{deep}\}.
\end{equation}

\subsubsection{Semantic-conditioned Cross-attention Fusion}
Then, we employ a semantic-conditioned cross-attention module to align acoustic and semantic features through an asymmetric attention mechanism. Rather than indiscriminately fusing both streams, we enable semantic representations to query and selectively attend to acoustic features, extracting evidence that is contextually relevant for depression detection.

Let $U^{\text{acoustic}} \in \mathbb{R}^{T \times d}$ and $U^{\text{semantic}} \in \mathbb{R}^{T \times d}$ denote the structured representations. Cross-level fusion is performed by treating deep semantic representations as queries and shallow acoustic representations as keys and values:
\begin{equation}
Q = W_Q U^{\text{semantic}}, \quad
K = W_K U^{\text{acoustic}}, \quad
V = W_V U^{\text{acoustic}},
\end{equation}
\begin{equation}
F = \text{softmax}\!\left(\frac{QK^\top}{\sqrt{d}}\right)V,
\end{equation}
where $W_Q$, $W_K$, and $W_V$ are learnable projection matrices. The fused representation $F$ is subsequently processed by a feed-forward network with layer normalization, followed by temporal average pooling to obtain utterance-level representations for classification, as shown in Table \ref{tab:classifier}.

\begin{table}[h]
    \centering
    \normalsize
    \caption{Architecture of the FFN block in Semantic-conditioned Cross-attention module.}
    \begin{tabular}{|c|c|}
        \hline
        \textbf{Layers} & \textbf{Output shape}\\
        \hline
        \hline
        Layer Normalization & (batch, 1024) \\
        Fully Connected (Linear) &  (batch, 64) \\
        SiLU Activation & (batch, 64) \\
        Dropout (0.3) & (batch, 64)\\
        Fully Connected (Linear) & (batch, 1024) \\
        \hline
    \end{tabular}
    \label{tab:classifier}
\end{table}

\subsubsection{Pseudo Token Target Generation}
\label{sec:pseudo_token}
To provide temporal structure constraints on depression-related features, we encourage them to be organized into a small number of stable temporal events.

\begin{figure}[h]
    \centering
    \includegraphics[width=0.6\linewidth]{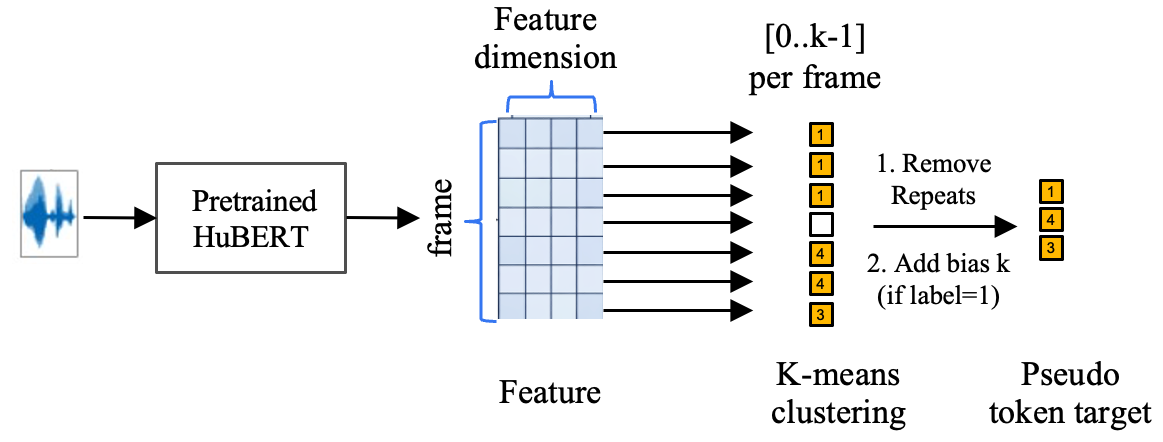}
    \caption{Flowchart of Pseudo Token Target Generation}
    \label{fig:CTC}
\end{figure}

As shown in Fig.~\ref{fig:CTC}, we first extract frame-level hidden representations of a given utterance from the 12th layer of a pretrained SSL model. For each utterance, we perform per-utterance $k$-means clustering over its frame-level embeddings to obtain a cluster assignment sequence $\{c_t\}_{t=1}^{T'}$, where $c_t \in \{0,\ldots,k-1\}$. Consecutive duplicate assignments are collapsed to form a compact discrete sequence.

To maintain separation between depression classes in the target space, we apply class-conditional label offsetting such that non-depressive (ND) utterances use labels $\{0,\ldots,k-1\}$, while depressive (D) utterances use $\{k,\ldots,2k-1\}$. Together with the pseudo token blank symbol at index $2k$, this yields a vocabulary of $2k+1$ output symbols. 

\subsection{Optimization Objective}

The proposed framework is trained with two complementary objectives. The primary objective is utterance-level binary classification for depression detection. The auxiliary CTC loss provides a sequential structure constraint that encourages the model to organize depression-related features into a small number of stable temporal events.

\paragraph{Utterance-Level Classification Loss}
We optimize utterance-level binary classification using Binary Cross-Entropy (BCE) loss:
\begin{equation}
\mathcal{L}_{\text{BCE}} = \frac{1}{N} \sum_{i=1}^{N} \left[ -y_i \log(p_i) - (1-y_i) \log(1-p_i) \right],
\end{equation}
where $N$ denotes the number of training utterances, $y_i \in \{0,1\}$ is the ground-truth label, and $p_i = \sigma(z_i)$ is the predicted probability from the classification head.

\paragraph{Temporal Structure Regularization}
We employ Connectionist Temporal Classification (CTC) loss \cite{graves2012connectionist} as an auxiliary objective to impose temporal structure constraints. Given the encoder output sequence and the pseudo token targets generated as described in Section~\ref{sec:pseudo_token}, the CTC loss is defined as:
\begin{equation}
\mathcal{L}_{\text{CTC}} = -\frac{1}{N} \sum_{i=1}^{N} \log P(Y_i | X_i),
\end{equation}
where $Y_i$ denotes the target sequence for utterance $i$, $X_i$ is the corresponding encoder output, and $P(Y_i | X_i)$ is the probability of all valid alignments between $X_i$ and $Y_i$ computed via the forward-backward algorithm.

The total training objective combines both losses: $\mathcal{L} = \mathcal{L}_{\text{BCE}} + \lambda \mathcal{L}_{\text{CTC}}$, where $\lambda$ is a weighting hyperparameter.

\section{EXPERIMENTAL SETUP}

To evaluate the effectiveness of the proposed \textit{HAREN}-CTC framework for depression detection, we conduct comprehensive experiments including comparison with state-of-the-art (SOTA) methods and ablation studies to quantify the contribution of individual components.

\subsection{Datasets and Preprocessing}

Experiments are conducted on two benchmark datasets: DAIC-WOZ~\cite{gratch2014distress} and MODMA~\cite{cai2022multi}.

The DAIC-WOZ dataset consists of multimodal clinical interviews collected from 189 English-speaking participants, including 56 subjects diagnosed with depression. Audio recordings range from 7 to 33 minutes and are sampled at 16 kHz. Following common practice, we segment participant utterances based on transcript timestamps and discard segments shorter than one second or dominated by silence.

To address class imbalance during training, we adopt the sampling strategy proposed in Speechformer~\cite{chen2022speechformer}, selecting the 18 longest utterances per non-depressed subject and 46 per depressed subject. During evaluation, the 20 longest utterances from each test subject are used. To increase temporal variability, a 10-second segment is randomly sampled from each utterance at every training epoch. We also correct an annotation error by updating the label of subject 409 from non-depressed to depressed.

The MODMA dataset contains interview recordings from 52 Mandarin-speaking participants, including 23 individuals diagnosed with depression and 29 healthy controls. Each recording comprises three tasks: free-form interview, text reading, and picture description. In this work, we use only the interview recordings to ensure consistency with DAIC-WOZ. We additionally correct a known labeling error by updating subject 2010037 from MDD to a healthy control.

Demographic and clinical statistics for both datasets are summarized in Tables~\ref{tab:DAICWOZ} and~\ref{tab:MODMA}.

\begin{table*}[htbp]
\caption{DAIC-WOZ: Gender Distribution, Diagnostic Category, and PHQ-8 Severity Scores (Mean ± SD)
\label{tab:DAICWOZ}}
\centering
\normalsize
\begin{tabular}{|c|c|c|c|c|c|c|c|}
\hline
\multirow{2}{*}{Gender} & \multirow{2}{*}{Category} & \multirow{2}{*}{Number} & \multicolumn{4}{c|}{PHQ-8 Score} & \multirow{2}{*}{PHQ Score Statistics} \\
\cline{4-7}
 & & & 0-4 & 5-9 & 10-19 & 20-24 & \\
\hline
\hline
Female & Control Group & 56 &38 & 18& 0& 0& 3.4±3.09 \\

Female & Depression Group & 31 & 0&0 & 25& 6& 14.5±4.19 \\

Male & Control Group & 76 & 48& 28& 0& 0& 6.8±5.92 \\

Male & Depression Group & 26 & 0& 0& 25& 1& 14.0±3.22 \\
\hline
\end{tabular}

\end{table*}

\begin{table*}[htbp]
\caption{MODMA: Gender Distribution, Diagnostic Category, and PHQ-9 Severity Scores (Mean ± SD) \label{tab:MODMA}}
\centering
\normalsize
\begin{tabular}{|c|c|c|c|c|c|c|c|}
\hline
\multirow{2}{*}{Gender} & \multirow{2}{*}{Category} & \multirow{2}{*}{Number} & \multicolumn{4}{c|}{PHQ-9 Score} & \multirow{2}{*}{PHQ Score Statistics} \\
\cline{4-7}
 & & & 0-4 & 5-9 & 10-19 & 20-27 & \\
\hline
\hline
Female & Control Group & 9 & 8& 1& 0& 0& 2.11±2.20 \\

Female & Depression Group & 7 & 0 &0  &3 & 4& 18.14±5.15 \\

Male & Control Group & 21 &18 & 3& 0& 0& 2.86±2.17 \\

Male & Depression Group & 15 & 0& 0& 10& 5& 18.87±2.92\\
\hline
\end{tabular}

\end{table*}

\subsection{Feature Extraction}

Raw audio signals are first preprocessed through an audio processing pipeline that includes random segmentation into 10-second segments, amplitude normalization to ensure consistent signal scales across samples, and padding with mask generation to handle variable lengths.

Then, we employ pretrained WavLM-Large~\cite{chen2022wavlm} as a frozen feature extractor, extracting hidden representations from all 24 transformer layers to preserve hierarchical acoustic and semantic information.

For auxiliary CTC supervision, HuBERT-Large~\cite{hsu2021hubert}, pretrained on Libri-Light and fine-tuned on LibriSpeech~\cite{panayotov2015librispeech}, is used to generate discrete pseudo-labels. All feature extraction procedures are implemented using the Fairseq toolkit~\cite{ott2019fairseq}.

\subsection{Implementation Details}

All experiments are implemented in PyTorch and conducted on a single NVIDIA V100 GPU. The batch size is set to 16 for DAIC-WOZ and 8 for MODMA. Optimization is performed using Adam with a weight decay of $1\times10^{-4}$. The learning rate is $1\times10^{-5}$ for DAIC-WOZ and $1\times10^{-4}$ for MODMA.

To further mitigate class imbalance, a weighted random sampler is employed during training.

During inference, predictions are generated at the utterance level. Subject-level predictions are obtained by aggregating utterance-level probabilities using confidence-weighted averaging. Model performance at the subject level is evaluated using Macro F1-score, Macro Recall or Unweighted Average Recall (UAR), and Macro Precision to account for class imbalance.

\subsection{Baseline Models}

we compare \textit{HAREN}-CTC with six representative baseline approaches, including:
\begin{itemize}
    \item DepAudioNet~\cite{ma2016depaudionet}: models long-term dependencies in raw audio by combining convolutional neural networks with long short-term memory and random sampling strategies.
    \item Speechformer~\cite{chen2022speechformer}: applies a hierarchical Transformer structure to model speech characteristics and structural information across frame, phoneme, word, and sentence levels.
    \item Vlad-GRU~\cite{shen2022automatic}: aggregates variable-length speech features using a NetVLAD layer combined with recurrent networks to capture depressive information.
    \item SFTN~\cite{han2024spatial}: extracts spatial spectral textures and temporal dynamics through parallel network branches to improve depression recognition performance.
    \item DALF~\cite{yang2023attention}: utilizes attention-guided learnable time-domain filterbanks instead of fixed filters to adaptively capture depression-sensitive frequency cues.
    \item DMFP~\cite{zhao2025decoupled}: decouples multi-perspective speech features to minimize redundancy and enhance representation capability before final fusion.
\end{itemize}

\subsection{Evaluation Protocols}

We evaluate \textit{HAREN}-CTC under two complementary evaluation settings, targeting both conventional benchmark comparability and realistic generalization ability.

\paragraph{Performance Upper-Bound Evaluation}
Following common practice in the AVEC challenge \cite{ringeval2017avec} and prior studies, this setting estimates the upper-bound performance under fixed official data splits. For DAIC-WOZ, we adopt the AVEC 2017 protocol \cite{ringeval2017avec} with 106 subjects for training, 35 for development. The best performance on the development set is selected for reporting. Evaluation metrics include Macro F1-score, Recall, and Precision, which are consistent with the official AVEC evaluation criteria and facilitate direct comparison with existing literature.

While this setting enables fair comparison with published results, selecting the best epoch on development set may introduce optimistic bias and does not fully reflect model robustness in real-world deployment scenarios.

\paragraph{Generalization Evaluation}
To rigorously evaluate model robustness and generalization, we merge all available data, including training, development, and test sets. Then we perform subject-level stratified 5-fold cross-validation to ensure no subject appears in multiple folds. This cross-validation procedure is repeated across 5 different random seeds, yielding 5 independent runs. For each run, we train models for a fixed number of epochs and average performance across the 5 folds. 

Performance is reported as the mean and standard deviation across the 5 independent runs using Macro F1-score, Macro Recall (UAR), Macro Precision, and AUC. To further evaluate the statistical significance of the proposed HAREN-CTC framework against baseline methods, we conducted significance tests across all performance metrics using paired t-tests \cite{hsu2014paired}. To further verify the reliability of the observed improvements, bootstrap resampling \cite{dixon2006bootstrap} with 1,000 iterations was applied to construct 95\% confidence intervals (CIs) for the performance differences. 

\subsection{Ablation Studies}

We conduct ablation studies to quantify the contribution of Hierarchical Acoustic Representation Encoder and Temporal structure regularization in \textit{HAREN}-CTC. All ablation experiments use identical data splits, preprocessing, and training configurations. Specifically, we compare the full model against two ablation baselines: (i) single-layer SSL baselines without hierarchical clustering or cross-level fusion, and (ii) a variant without the auxiliary CTC supervision.

\section{EXPERIMENTAL RESULTS \& ANALYSIS}

\begin{table*}[htbp]
\caption{Performance comparison under the performance upper-bound evaluation. Results in the first row are reproduced based on DepAudioNet \cite{ma2016depaudionet}. }
\centering
\normalsize
\resizebox{0.58\textwidth}{!}{  
\begin{tabular}{|c|c|c|c|c|c|}
\hline
Datasets & Methods & Features & Macro F1 & Recall & Precision\\
\hline
\hline
\multirow{9}{*}{DAIC-WOZ} & DepAudioNet \cite{ma2016depaudionet} & MFbanks & 0.61 & 0.77 & 0.68  \\
& Speechformer \cite{chen2022speechformer}& HuBERT & 0.69 & - & - \\
& MSCDR \cite{du2023depression} & LPC-MFCC & 0.75 & 0.75& 0.75\\
& Vlad-GRU \cite{shen2022automatic} & MFbanks & 0.77 & \textbf{1.00} & 0.63 \\
& SFTN \cite{han2024spatial}& MFbanks & 0.76 & 0.92 & 0.65 \\
& DALF \cite{yang2023attention} & RS & 0.78 & 0.79 & 0.77 \\
& DMPF \cite{zhao2025decoupled}& RS &0.80 & 0.81 & 0.80\\
& \textbf{\textit{HAREN}-CTC} & WavLM &\textbf{0.81} &0.81& \textbf{0.81} \\
\hline
\hline
\multirow{4}{*}{MODMA} &DepAudioNet \cite{ma2016depaudionet} &  MFbanks & 0.62 & 0.56 &  0.77\\
&Vlad-GRU \cite{shen2022automatic} & MFbanks  & 0.60 &0.67&0.55 \\
&DMPF \cite{zhao2025decoupled}& RS & 0.76 & 0.77 & 0.76  \\ 
&\textbf{\textit{HAREN}-CTC} & WavLM & \textbf{0.82} & \textbf{0.83}& \textbf{0.86}\\
\hline
\end{tabular}
}
\label{tab:result1}
\end{table*}

\begin{table*}[t]
\caption{Performance comparison under the generalization evaluation scenario. Results in the first row are reproduced from DepAudioNet \cite{ma2016depaudionet}. Speechformer \cite{chen2022speechformer} requires utterance-level timestamps, which are not available in the MODMA dataset; therefore, we do not include Speechformer results for MODMA.}
\centering
\normalsize
\resizebox{0.75\textwidth}{!}{
\begin{tabular}{|c|c|c|c|c|c|}
\hline
Datasets & Methods &  Macro F1 & UAR & Macro Precision & AUC\\
\hline
\hline
\multirow{3}{*}{DAIC-WOZ} & DepAudioNet \cite{ma2016depaudionet} & 0.533 (0.03)  &  0.562 (0.03) & 0.508 (0.03) & 0.566 (0.04)  \\
& Speechformer \cite{chen2022speechformer} & 0.533 (0.03) & 0.538 (0.02) & 0.539 (0.03) & 0.557 (0.01)\\
& \textbf{\textit{HAREN}-CTC} &\textbf{0.576 (0.02)} & \textbf{0.592 (0.03)}&  \textbf{0.599 (0.02)} & \textbf{0.599 (0.02)}\\
\hline
\hline
\multirow{2}{*}{MODMA} &DepAudioNet \cite{ma2016depaudionet}  & 0.557 (0.03) & 0.562 (0.03) &  0.566 (0.03)&  0.586 (0.03)\\
&\textbf{\textit{HAREN}-CTC}  & \textbf{0.644 (0.03)}& \textbf{0.665 (0.02)}& \textbf{0.680 (0.03)} & \textbf{0.680 (0.03)}\\
\hline

\end{tabular}
}
\label{tab:genresult}
\end{table*}

\subsection{Comparison to the State-of-the-Art}
This section present the results under two evaluation settings. Quantitative results are summarized in Tables~\ref{tab:result1} and~\ref{tab:genresult}, respectively.

\paragraph{Performance Upper-Bound Evaluation}

\begin{figure}
    \centering
    \includegraphics[width=0.5\linewidth]{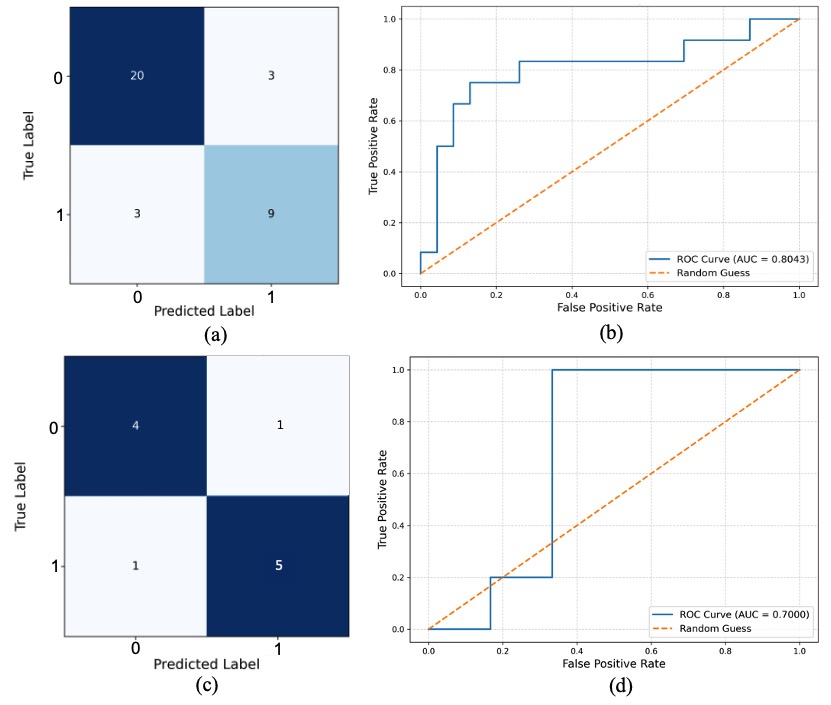}
    \caption{Confusion matrix and ROC AUC Curve on the development set of the DAIC-WOZ dataset in performance upper-bound evaluation; (a) Confusion Matrix, (b) ROC AUC Curve.}
    \label{fig:conf_daic}
\end{figure}

Table~\ref{tab:result1} reports results under the performance upper-bound evaluation. Figure \ref{fig:conf_daic} shows the confusion matrix and ROC AUC Curve on the development set of the DAIC-WOZ dataset in performance upper-bound evaluation. \textit{HAREN}-CTC achieves a Macro F1 score of 0.81 on DAIC-WOZ and 0.82 on MODMA, outperforming all baseline methods on both datasets. On MODMA, our method further demonstrates strong performance with a UAR of 0.83 and Macro Precision of 0.86. These results validate the effectiveness of the proposed framework under standard experimental settings. However, since this evaluation relies on fixed splits and development-set-based model selection, the reported performance serves as an optimistic upper bound rather than a rigorous assessment of generalization.

\paragraph{Generalization Evaluation}
Table~\ref{tab:genresult} reports results under the generalization evaluation, which provides a more rigorous assessment of model robustness under realistic deployment conditions. On DAIC-WOZ, \textit{HAREN}-CTC consistently outperforms DepAudioNet~\cite{ma2016depaudionet} and Speechformer~\cite{chen2022speechformer} across all metrics (Macro F1, UAR, Macro Precision, and AUC) with smaller performance variance across folds and seeds. This improved stability indicates reduced sensitivity to subject partitioning, a critical requirement for real-world screening scenarios.

As shown in Table \ref{significance_clean}, paired t-test results demonstrate statistical significance for Macro Precision and AUC in Speechformer \cite{chen2022speechformer}, with bootstrap confidence intervals of [0.004, 0.117] and [0.012, 0.082] confirming these improvements' reliability. When applied to the DepAudioNet \cite{ma2016depaudionet}, Macro Precision shows particularly noteworthy improvement, achieving statistical significance. This substantial gain suggests that \textit{HAREN}-CTC effectively reduces false positives—critical in clinical depression detection. The bootstrap confidence intervals further validate these findings, with several key metrics showing intervals excluding zero.

On MODMA, \textit{HAREN}-CTC demonstrates even larger improvements over baselines. Notably, the performance gap between upper-bound and generalization settings differs across datasets. DAIC-WOZ exhibits a substantial drop in Macro F1 score from 0.81 to 0.576 (approximately 28.5\%), while MODMA shows smaller degradation from 0.82 to 0.644 (approximately 21\%). This difference may be attributed to dataset characteristics: DAIC-WOZ features human-computer interviews in English, whereas MODMA consists of human-to-human clinical conversations in Mandarin. Depression-related acoustic patterns may manifest differently across languages, conversational styles, and cultural norms, affecting how well decision boundaries generalize across subject partitions. 

Despite these cross-linguistic and cross-cultural variations, \textit{HAREN}-CTC achieves consistent improvements on both datasets, suggesting the model captures more invariant depression-relevant patterns and is less dependent on dataset-specific biases.

\begin{table*}[htbp]
\caption{Significance tests of \textit{HAREN}-CTC against baselines. $\Delta=\mu_{\text{HAREN-CTC}}-\mu_{\text{Baseline}}$; Rel.\ denotes relative improvement over the baseline mean. ${}^{*}$ indicates paired $t$-test $p<0.05$.}
\centering
\normalsize
\setlength{\tabcolsep}{5pt}
\renewcommand{\arraystretch}{1.18}

\resizebox{1.0\textwidth}{!}{%
\begin{tabular}{|c|c|c|c|
                >{\centering\arraybackslash}p{5.2cm}|
                >{\centering\arraybackslash}p{4.0cm}|}
\hline
\multirow{2}{*}{Baselines} & \multirow{2}{*}{Metrics} & \multirow{2}{*}{$\Delta$} & \multirow{2}{*}{Rel.(\%)} &
\multicolumn{2}{c|}{Significance Tests} \\
\cline{5-6}
 &  &  &  &
\makecell{Paired $t$-test\\ $t$ / $p$ / 95\% CI} &
\makecell{Bootstrap\\ $p$ / 95\% CI} \\
\hline
\hline

\multirow{4}{*}{Speechformer}
& Macro F1        & 0.043 & 8.0  & $2.19$ / $0.094$  / $[-0.011,\,0.097]$ & \textbf{$0.0082$} / $[0.009,\,0.076]$ \\
& UAR             & 0.054 & 10.1 & $2.70$ / $0.054$  / $[-0.002,\,0.110]$ & \textbf{$0.0054$} / $[0.016,\,0.083]$ \\
& Macro Precision & 0.060 & 11.2 & $2.98$ / \textbf{$0.0407$}${}^{*}$ / $[0.004,\,0.117]$ & \textbf{$0.0008$} / $[0.024,\,0.092]$ \\
& AUC             & 0.047 & 8.4  & $3.71$ / \textbf{$0.0207$}${}^{*}$ / $[0.012,\,0.082]$ & \textbf{$<10^{-4}$} / $[0.024,\,0.067]$ \\
\hline
\hline

\multirow{4}{*}{DepAudioNet}
& Macro F1        & 0.043 & 8.0  & $2.64$ / $0.058$ / $[-0.002,\,0.088]$ & \textbf{$0.0012$} / $[0.014,\,0.070]$ \\
& UAR             & 0.030 & 5.4  & $2.33$ / $0.080$ / $[-0.006,\,0.066]$ & \textbf{$<10^{-4}$} / $[0.011,\,0.055]$ \\
& Macro Precision & 0.091 & 17.9 & $4.45$ / \textbf{$0.0112$}${}^{*}$ / $[0.034,\,0.148]$ & \textbf{$<10^{-4}$} / $[0.053,\,0.124]$ \\
& AUC             & 0.038 & 6.7  & $1.63$ / $0.179$ / $[-0.027,\,0.102]$ & \textbf{$<10^{-4}$} / $[0.011,\,0.085]$ \\
\hline

\end{tabular}%
}
\label{tab:significance_clean}
\end{table*}

\begin{table}[t]
\caption{Ablation study results on the DAIC-WOZ dataset under two settings.}
\centering
\normalsize
\resizebox{0.58\textwidth}{!}{
\begin{tabular}{|c|c|c|c|}
\hline
Settings & \textit{HAREN} & CTC-branch &M-F1 \\
\hline
\hline
\multirow{3}{*}{Performance upper-bound evaluation} & \checkmark & \checkmark &\textbf{0.82}\\
& \ding{56} & \checkmark &0.71\\
& \checkmark& \ding{56} & 0.77\\
\hline
\hline
\multirow{3}{*}{Generalization evaluation} &\checkmark & \checkmark &\textbf{0.58}\\
& \ding{56} & \checkmark& 0.50 \\
& \checkmark& \ding{56} & 0.52\\

\hline
\end{tabular}
}
\label{tab:ablation}
\end{table}

\subsection{Ablation Study}
\subsubsection{Component-wise Ablation Analysis}
To quantify the individual contributions of the proposed components, we conduct ablation studies by removing the hierarchical adaptive representation encoder (HAREN) or the CTC-based temporal regularization. Table~\ref{tab:ablation} reports results under both evaluation settings on DAIC-WOZ.

Under the performance upper-bound evaluation, removing HAREN decreases Macro F1 from 0.82 to 0.71, demonstrating the importance of hierarchical feature disentanglement and semantic-conditioned fusion. Removing the CTC branch results in a smaller drop to 0.77, indicating that temporal structure regularization provides complementary benefits. Under the more rigorous generalization evaluation, the contribution of each component becomes more pronounced. Without HAREN, performance drops from 0.58 to 0.50, while removing CTC reduces performance to 0.52. These results indicate that both components are essential for robust generalization, with HAREN playing a more critical role in capturing depression-relevant patterns that transfer across subject partitions. The larger performance gap in the generalization setting suggests that hierarchical representation learning and temporal regularization are particularly valuable for mitigating overfitting to subject-specific characteristics.

\begin{figure}[t]
    \centering
    \includegraphics[width=0.7\linewidth]{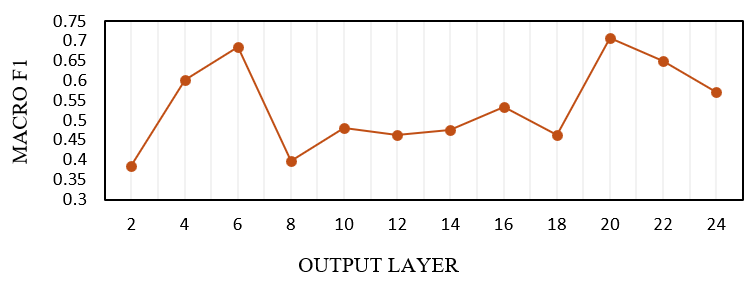}
    \caption{Comparison of the model's performance without \textit{HAREN} on the DAIC-WOZ development set using representations from different hidden layers of the pretrained WavLM.}
    \label{fig:M1}
\end{figure}

\begin{figure}[t]
    \centering
    \includegraphics[width=0.5\linewidth]{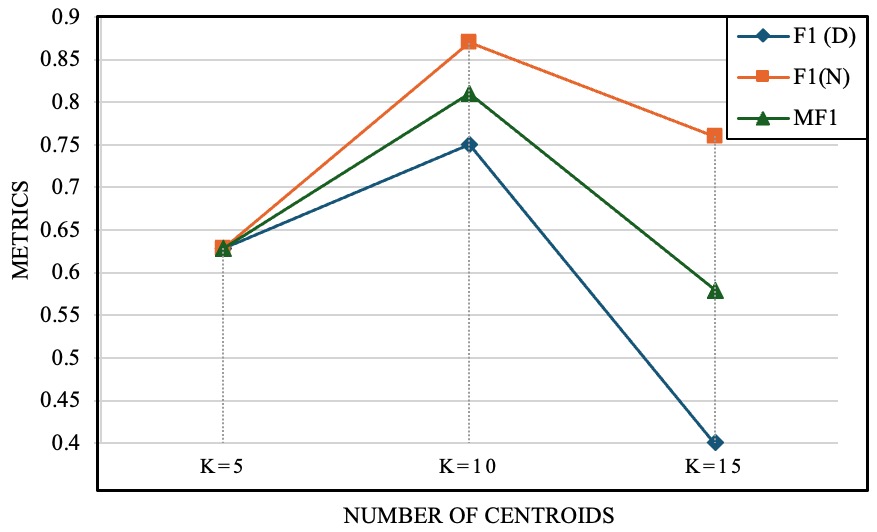}
    \caption{Comparison of Performance Metrics for Different Numbers of Centroids (K = 5, 10, 15) on the DAIC-WOZ development set.}
    \label{fig:M4}
\end{figure}

\subsubsection{Layer-wise Representation Analysis}
To further examine the limitations of fixed-layer representations, we conduct a layer-wise performance analysis using single-layer baselines without HAREN. As shown in Figure~\ref{fig:M1}, performance exhibits substantial variation across layers, with Macro F1 scores ranging from 0.38 to 0.70. Although certain layers like layers 6 and 20 achieve relatively higher scores, sharp performance degradation occurs at others like layers 2, 8, and 18, revealing strong sensitivity to representation depth. Critically, even the best-performing single layer falls short of the full \textit{HAREN}-CTC model, suggesting that no individual layer encodes sufficient depression-relevant information. These findings indicate that HAREN effectively mitigates layer-specific sensitivity by integrating representations across multiple depths, thereby enabling the model to jointly capture fine-grained prosodic patterns and high-level semantic features that are collectively indicative of depression.

\subsubsection{CTC Clustering Resolution Analysis}
As shown in Figure \ref{fig:M4}, performance peaks when $K=10$, indicating that this clustering resolution strikes the best balance between temporal granularity and label consistency. Lower ($K=5$) and higher ($K=15$) values lead to under-segmentation and over-fragmentation, respectively, both of which hurt performance. These results underscore the importance of tuning the CTC clustering resolution for optimal performance.

\section{CONCLUSION \& FUTURE WORK}

This work presents \textit{HAREN}-CTC, a deep learning framework for speech-based depression detection that explicitly aligns shallow acoustic and deep semantic representations from pretrained self-supervised learning (SSL) models. Unlike approaches relying on single-layer SSL features or ad hoc fusion, \textit{HAREN}-CTC introduces an inductive bias that structures hierarchical representations, allowing fine-grained acoustic variations to be interpreted within semantic context. Additionally, we propose an alignment-aware Connectionist Temporal Classification (CTC) objective as auxiliary supervision to address the sparse, irregular temporal distribution of depressive speech cues without frame-level annotations.

Extensive experiments on DAIC-WOZ and MODMA benchmarks show that our framework consistently outperforms strong single-layer SSL baselines and achieves state-of-the-art results under both upper-bound and generalization settings. Ablation studies confirm that explicitly structuring hierarchical representations and modeling their interactions are critical for capturing depression-relevant patterns and improving robustness.

Two promising directions emerge for future work. First, while our framework performs well under limited data, data generation and augmentation strategies could further enrich training distributions. Generating samples for underrepresented classes and diversifying speaking styles, recording conditions, and demographics may improve robustness without requiring large-scale clinical data collection.

Second, the current utterance-level modeling with subject-level aggregation could be extended to long-context modeling at the session or dialogue level. Incorporating hierarchical conversation encoders or long-range sequence modeling would enable the system to capture emotion trajectories and interaction patterns across entire interviews, revealing temporal dynamics that complement acoustic cues.

\section{Data Availability}
The DAIC-WOZ dataset is publicly available at (https://dcapswoz.ict.usc.edu/).

The MODMA dataset is publicly available at (https://modma.lzu.edu.cn/data/index/).

\section*{Acknowledgments}
This work was supported by the Lien Foundation, Singapore.


\bibliographystyle{unsrt}  
\bibliography{references}

\end{document}